\documentclass{Interspeech2024}

\usepackage[T1]{fontenc}
\usepackage[utf8]{inputenc}

\interspeechcameraready

\title{On the Effect of Purely Synthetic Training Data\\for Different Automatic Speech Recognition Architectures}

\name[affiliation={1,2,*}]{Benedikt}{Hilmes}
\name[affiliation={1,2,*}]{Nick}{Rossenbach}
\name[affiliation={1,2}]{Ralf}{Schlüter}

\address{
  $^1$RWTH Aachen University, Germany\\
  $^2$AppTek GmbH, Germany\\
  $^*$\textit{\small equal contribution}
 }
\email{<lastname>@ml.rwth-aachen.de}


\newcommand{\alert}[1]{{#1}}

\usepackage{amsmath,graphicx}
\usepackage{multirow}
\usepackage{glossaries} 
\usepackage{makecell}
\usepackage{multicol}
\usepackage{tikz}
\usepackage{hyperref}
\usetikzlibrary{arrows,positioning} 
\usetikzlibrary{shapes,decorations}
\usetikzlibrary{calc}
\usetikzlibrary{decorations.pathreplacing}
\usepackage{cleveref}
\usepackage{xcolor}
\usepackage{adjustbox}
\usetikzlibrary{arrows,positioning} 
\usetikzlibrary{shapes,decorations}
\usetikzlibrary{calc}
\usetikzlibrary{decorations.pathreplacing}
\definecolor{blind_red}{HTML}{D7191C}
\definecolor{blind_orange}{HTML}{FDAE61}
\definecolor{blind_yellow}{HTML}{FFFFBF}
\definecolor{blind_blue}{HTML}{ABD9E9}
\definecolor{blind_blue2}{HTML}{2C7BB6}

\newacronym{TTS}{TTS}{text-to-speech}
\newacronym{ASR}{ASR}{automatic speech recognition}
\newacronym{LSTM}{BiLSTM}{bi-directional LSTM}
\newacronym{GL}{G\&L}{Griffin \& Lim}
\newacronym{GMM}{GMM}{Gaussian mixture model}
\newacronym{NAR}{NAR}{non-autoregressive}
\newacronym{HMM}{GMM-HMM}{Gaussian mixture based system using hidden Markov model}
\newacronym{WER}{WER}{word error rate}
\newacronym{MFA}{MFA}{Montreal forced aligner}
\newacronym{VTLN}{VTLN}{vocal tract length normalization}
\newacronym{SAT}{SAT}{speaker adaptive training}
\newacronym{NN}{NN}{neural network}
\newacronym{AED}{AED}{attention-based encoder-decoder}
\newacronym{LM}{LM}{language model}
\newacronym{CV}{CV}{cross validation}
\newacronym{Hybrid}{Hybrid}{hybrid deep neural network HMM}

\begin{document}

\maketitle

\begin{abstract}
\alert{
In this work we evaluate the utility of synthetic data for training automatic speech recognition (ASR).
We use the ASR training data to train a text-to-speech (TTS) system similar to FastSpeech-2.
With this TTS we reproduce the original training data, training ASR systems solely on synthetic data.
For ASR, we use three different architectures, attention-based encoder-decoder,  hybrid deep neural network hidden Markov model and a Gaussian mixture hidden Markov model, showing the different sensitivity of the models to synthetic data generation.
In order to extend previous work, we present a number of ablation studies on the effectiveness of synthetic vs. real training data for ASR.
In particular we focus on how the gap between training on synthetic and real data changes by varying the speaker embedding or by scaling the model size.
For the latter we show that the TTS models generalize well, even when training scores indicate overfitting.
}
\end{abstract}

\section{Introduction}
Current literature shows the capability of synthetic data to complement real data and thus improve \gls{ASR} training through various ways and techniques \cite{Laptev-2020-YouDoNotNeedMore, 9688255, 8683480, Baskar-2021-EatEnhancedASR-TT, Hu-2022-SYNTUtilizingIm} .
Commonly end-to-end architectures are trained with a combination of real and synthetic data, where especially models like the \gls{AED} benefit from the additional synthetic data \cite{9688255, Baskar-2021-EatEnhancedASR-TT, Hu-2022-SYNTUtilizingIm}.
\alert{However, we still lack a good understanding of how well synthetic data is able to replace real data. 
To this end, we suggest to use synthetic training data \textit{only} to analyze and compare its \gls{ASR} training utility against real data.
Such a study helps to gain further insight on the discrepancy between synthetic and real data. 
Recent work has presented large TTS systems trained on much more data than typically available for academically defined tasks \cite[p. 5]{anastassiou2024seedtts}.
But even with such an industrial scale system it was not possible to create synthetic data that is equally utilizable to real data.}
In this work we explore the \alert{corresponding} performance gap for three \gls{ASR} architectures: A classic \gls{HMM}, a \gls{Hybrid} \cite{bourlard2012connectionist} and an \gls{AED} \cite{7472621} system.
\alert{A lot of previous work on synthetic data analysis was done on data generated by autoregressive \gls{TTS} models, like Tacotron-2 \cite{Shen-2018-NaturalTTSSynthesi}, with the exception of \cite{9688218}.}
In this work we focus on a simple non-autoregressive TTS model which is similar to FastSpeech, but with a BLSTM-decoder, and an encoder with mixed convolutions and BLSTM.
\alert{There} are more recent architectures which exhibit strong performance on standard TTS tasks.
\alert{However,} we consider a more simple and established TTS architecture that we know from previous experiments \cite{10389782} to work on ASR-specific training data \alert{and thus is expected} to be better suited for our analysis.
The contributions of this work are as follows:

\begin{itemize}
\item Investigating how robust traditional ASR approaches such as \gls{HMM} react to synthetic data compared to a more modern \gls{Hybrid} or AED system.

\item Quantifying the impact of low-quality Griffin \& Lim vocoding for the usability of audio data.

\item Showing how simply increasing the \alert{number of TTS} model parameters already improves the usability of the synthetic data for ASR.

\item Showing that in the context of this work, more sophisticated speaker embedding systems greatly influence the performance.

\end{itemize}
Especially\alert{,} the effect of hyper-parameter tuning \alert{rarely} is covered in TTS-literature due to the expensive  MOS evaluations \alert{this} would \alert{ensue}.
A large scale comparison of training various ASR architectures with both real and synthetic data has been done in \cite{9688255}.
\alert{We} extend \cite{9688255} and \cite{10389782} through \alert{analyzing the effect of ASR performance by using synthetic data \textit{only} for ASR training}.
Our work and software is publicly accessible and will remain as such \footnote{\url{https://github.com/rwth-i6/returnn-experiments/tree/master/2024-pure-synthetic-data}}.
\begin{table*}[t]
	\caption[Baseline results]{\alert{E}valuation on LibriSpeech \textit{dev-clean} and \textit{dev-other} corpora. \alert{Only data from LibriSpeech train-clean-100 is used for TTS training.} \textit{Vocoding only} used features extracted from the real data, vocoded by Griffin-Lim. \gls{TTS}-Durations indicates whether the model predicts phoneme durations via the duration predictor or is fed the ground truth durations from the alignment.}
		\begin{center}
			\begin{tabular}{| c | c | c | c | c | c | c | c | c | }
				\hline
				\multirow{3}{*}{Data}& \multirow{3}{*}{\shortstack[c]{Silence \\ Removal}} & \multirow{3}{*}{\shortstack[c]{TTS- \\ Durations}} & \multirow{3}{*}{\shortstack[c]{Data \\ Length}} &\multicolumn{5}{c|}{\alert{WER} [\%]}\\
				\cline{5-9}
				& & & &\multicolumn{2}{c}{\gls{HMM}} & \multicolumn{2}{|c|}{\gls{AED}} & \gls{Hybrid} \\
				\cline{5-9}
				&  &  & & clean & other & clean & other & other \\
				\hline
				\hline
				\multirow{2}{*}{Real} & No & - & 100.6h & \textbf{\phantom{0}8.1} &\textbf{25.9} &\textbf{\phantom{0}7.5} &\textbf{18.9}&\textbf{15.0}\\
				\cline{2-9}
				&Yes & - &\phantom{0}88.7h &\phantom{0}8.7 &28.0 & \phantom{0}7.8 & 20.3 & 15.5\\
				\hline
				\hline
				Vocoding & No & - & 100.6h &\phantom{0}8.7 & 27.7 & \phantom{0}7.6 & 19.4 & 15.2\\
				\cline{2-9}
				Only & Yes & - &\phantom{0}88.7h &  \phantom{0}9.5 & 28.8 & \phantom{0}8.0 & 20.1 & 16.1\\
				\hline
				\hline
				\multirow{2}{*}{Synthetic}& \multirow{2}{*}{Yes} & pred. &\phantom{0}81.1h & 10.0 & 32.2 & 14.1 & 37.6 & 26.2 \\
				\cline{3-9}
				&  & real & \phantom{0}88.3h& \phantom{0}9.7 & 32.2 & 10.9 & 31.8 & 26.8\\
				\hline
			\end{tabular}
		\end{center}
	\label{tab:baseline}
\end{table*}

\vspace{-0.25cm}
\section{Speech Synthesis}

As \gls{TTS} system we use the \gls{NAR} model from \cite{10389782}, which is closely related to \cite{Perez-Gonzalez-de-Martos-2021-VRAIN-UPVMLLPssys}, extended with Gaussian Upsampling \cite{Shen-2020-Non-AttentiveTacotr}.
It consists of a phoneme encoder, duration predictor and a feature decoder, which follow exactly the design as in \cite{10389782}.
For both duration prediction and decoding we pass a speaker embedding \alert{to enable multi speaker capabilities}.
In the simplest case this is generated by look-up table on the speaker ID which is learned during training, but we also investigate different more elaborate approaches, by generating fixed speaker embeddings through stronger pre-trained models.
Training is done on the reference durations, extracted from a given alignment. 
As spectrogram targets we use globally normalized 80-dimensional log-mel features with frame shift 12.5 ms and window size of 50 ms.
The spectrogram predictions are transformed into 512-dimensional linear features for \gls{GL} vocoding \cite{DBLP:conf/icassp/GriffinDL84} via a pre-trained \gls{LSTM} network. 
\alert{As validated in the baseline experiments of this paper, simple \gls{GL} vocoding does not reduce the quality of the generated audio for \gls{ASR} training.}
Counting both neural models, our architecture consists of 63M parameters.
\alert{The phoneme set consists of ARPA-BET phoneme symbols without stress marker. We mark word boundaries and possible silence with a \texttt{[space]} token between the last phoneme of a word and the first of the next respectively.}
 
\section{Speech Recognition}

Our \textbf{\gls{HMM}} model is implemented in RASR \cite{Wiesler-2014-RASRNNTheRWTHne}, which is functionally close to the \gls{MFA} commonly used for \gls{TTS} works such as FastSpeech-2 \cite{Ren-2020-FastSpeech2Fasta}. Training parameters are optimized on LibriSpeech-100h with focus on \gls{ASR} performance. The overall training consists of the several steps, which in more detail are explained in \cite{10389782}.
In this work we use the \gls{HMM} twofold. First we use the alignments produced by the system as ground truth alignments for duration prediction in our \gls{TTS}. For this we calculate the Viterbi path for our best alignment, setting a duration of zero for \texttt{[space]} tokens where no silence was aligned. For recognition, we use the model together with a pre-trained 4-gram count-based \gls{LM} from the LibriSpeech \cite{librispeech} dataset.

On top of the \gls{HMM} we use a \textbf{\gls{Hybrid}} model which predicts the frame-label posterior probability by a neural network. We use the final alignment output from the \gls{ASR} \gls{HMM} as training targets. 
The \gls{NN} consists of a stack of 8 1024-dimensional \gls{LSTM} layers followed by a linear layer with softmax activation and output size 12001 to match the corresponding CART \cite{CART} labels. \alert{In total the model consists of 210M parameters}. For recognition we again use the LibriSpeech 4-gram \gls{LM}.

Our last model is an \textbf{\gls{AED}} model as used in \cite{10389782} with 12 conformer blocks as encoder and single layer LSTM for the decoder, \alert{resulting in 98M parameters}. We use BPE labels \cite{sennrich-etal-2016-neural} with 2k merge operations as the output units.
Different from the \gls{TTS} model we use a frame shift to 10 ms and the window size to 25 ms for the feature extraction. The models uses a downsampling factor of 6. 
\alert{To improve training stability we increase the number of encoder layers over time, starting with 2 layers which are increased by 2 every 5 sub-epochs beginning the first increase after 10 sub-epochs reaching full model size at 10 full epochs (30 sub-epochs).}
For data augmentation we use SpecAgument \cite{park19e_interspeech} and apply speed-pertubation via librosa.resample()\footnote{\url{https://librosa.org/doc/latest/generated/librosa.resample.html}} uniformly distributing scales 0.9/1.0/1.1 among the input. 
Recognition results are without the use of an external language model.

\begin{table*}[h]
	\caption[Generalization Results]{Generalization Results. \alert{E}valuation of LibriSpeech \textit{dev-clean} and \textit{dev-other} corpora. \alert{Only data from LibriSpeech train-clean-100 is used for TTS training. The synthetic data is created by using either text from train-clean-100h (LS-100) or an equivalent subset of train-clean-360h (100h-LS-360). $\gamma$ denotes scaling the hidden dimension of the TTS by this factor. $\omega$ denotes scaling the number of layers by this factor.}}	
	\vspace{-1.0em}
	\begin{center}
		\begin{tabular}{| c | c | c | c | c | c | c | c | c | c |}
			\hline
			\multirow{3}{*}{Data}& \multirow{3}{*}{\shortstack[c]{\alert{(synth.)} \\ \alert{Audio} \\ \alert{Data}}}& \multirow{3}{*}{\shortstack[c]{TTS- \\ Dur.}} & \multirow{3}{*}{\shortstack[c]{Scale \\Type}} & \multirow{3}{*}{\shortstack[c]{Model \\ Scale}} & \multicolumn{5}{c|}{\alert{WER [\%]}}\\
			\cline{6-10}
			& & & & & \multicolumn{2}{c|}{\gls{HMM}} & \multicolumn{2}{c|}{\gls{AED}} & \gls{Hybrid} \\
			\cline{6-10}
			& & & & & clean & other & clean & other & other\\
			\hline
			\hline
			Real & LS-100 & - & - & - & \textbf{\phantom{0}8.1} &\textbf{25.9} &\textbf{\phantom{0}7.5} &\textbf{18.9}&\textbf{15.0}\\
			\hline
			\hline			
			\multirow{12}{*}{\alert{Synth.}}& \multirow{8}{*}{\alert{LS-100}}& \multirow{4}{*}{\alert{pred.}} & \multirow{1}{*}{-} & \multirow{1}{*}{-} & 10.0 & 32.2 & 14.1 & 37.6 & 26.2 \\
			\cline{4-10}	
			& & &\multirow{2}{*}{Dimension} & \multirow{1}{*}{$\gamma = 1.5$} & \phantom{0}9.8 & 31.3 & 11.7 & 32.4 & 23.4 \\				
			\cline{5-10}			
			& & & &\multirow{1}{*}{$\gamma = 2.0$} & \textbf{\phantom{0}9.5} & \textbf{31.0} & \textbf{10.5} & \textbf{30.5} & \textbf{22.1}\\								
			\cline{4-10}	
			& & &\multirow{1}{*}{Layers} &\multirow{1}{*}{$\omega = 2\phantom{.0}$} &\phantom{0}9.7 & 31.8 & 13.5 & 36.1 & 24.4\\
\cline{3-10}
			& &\multirow{4}{*}{\alert{real}} & \multirow{1}{*}{-} & \multirow{1}{*}{-} & \phantom{0}9.7 & 32.2 & 10.9 & 31.8 & 26.8\\	
			\cline{4-10}			
			& & &\multirow{2}{*}{Dimension} & \multirow{1}{*}{$\gamma = 1.5$} & \phantom{0}9.6 & 30.7 & \phantom{0}9.7 & 28.9 & 24.5\\			
			\cline{5-10}		
			& & & & \multirow{1}{*}{$\gamma = 2.0$} & \phantom{0}9.4 & \textbf{30.5} & \textbf{\phantom{0}9.3} & \textbf{27.6} & \textbf{23.1}\\			
			\cline{4-10}			
			& & & \multirow{1}{*}{Layers} &\multirow{1}{*}{$\omega = 2\phantom{.0}$} &\textbf{\phantom{0}9.2} & 30.9 & 10.2 & 29.9 & 24.8\\		
			\cline{2-10}
			& \multirow{4}{*}{\shortstack[c]{100h- \\ LS-360}} &\multirow{4}{*}{\alert{pred.}}& - & - & 10.3 & 32.9 & 19.1 & 43.8 & 26.9 \\
			\cline{4-10}
			&& & \multirow{2}{*}{Dimension} & $\gamma = 1.5$ &\phantom{0}9.9 & 32.9 & 16.1 & 38.0 & 24.5\\			
			\cline{5-10}			
			&& & &$\gamma = 2.0$ &\textbf{\phantom{0}9.8} & \textbf{31.3} & \textbf{15.0} & \textbf{35.7} &\textbf{23.2}\\
			\cline{4-10}
			& & & Layers & $\omega = 2\phantom{.0}$& 10.1 & 32.2 & 18.6 & 41.7 & 24.7 \\
			\hline			
	\end{tabular}
	\end{center}
	\label{tab:general}
\end{table*}

\section{Pipeline}

\begin{figure}
	\begin{center}
		\resizebox{3.0in}{!}{
			\begin{tikzpicture}[auto]
				\node[rectangle, minimum size=1.75cm, draw, fill=blind_blue!80, align=center] (gmm) at (0, 0) {Aligner \\ training};
				\node[rectangle, minimum size=1.75cm,draw, fill=blind_blue!80, right=20 pt of gmm, align=center] (align) {Forced \\ align};
				\node[rectangle, minimum size=1.75cm,draw,  fill=blind_orange!80, right=20 pt of align, align=center] (tts) {TTS \\ training};
				\node[rectangle, minimum size=1.75cm,draw,  fill=blind_orange!80, right=20 pt of tts, align=center] (synth) {Synthesis};
				\node[rectangle, minimum size=1.75cm,draw,  fill=blind_red!75, right=20 pt of synth,align=center] (asr) {ASR \\ training};
				\node[below= 18 pt of synth,align=center] (real) {random \\ speaker tag};
				\node[above= 18 pt of gmm] (data) {pre-processed data};
				\draw [->, line width=0.5mm] (gmm) -- (align);
				\draw [->, line width=0.5mm] (align) -- (tts);
				\draw [->, line width=0.5mm] (tts) -- (synth);
				\draw [->, line width=0.5mm] (synth) -- (asr);
				\path (align.south east) edge[bend right,->, line width=0.5mm, dashed] node [below=4pt, align=center] {oracle \\ durations} (synth.south west);
				\draw [->, dashed, line width=0.5mm] (real) -- (synth);
				\draw [->, line width=0.5mm] (data) -- (gmm);
				\path (data.east) edge[->, bend left = 18, line width=0.3mm] (tts.north west);				
			\end{tikzpicture}
		}
	\end{center}
	\vspace{-1.5em}
	\caption[Experiment pipeline]{Experiment pipeline for synthetic data training.}
	\label{fig:pipeline}
	\vspace{-0.5em}
\end{figure}
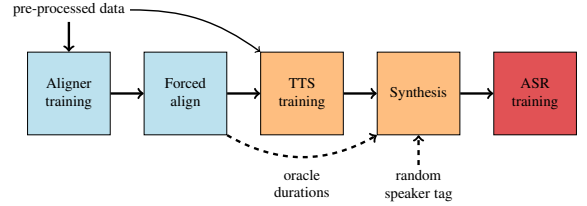

In general our pipeline consists of 5 steps analogue to \cite{10389782}, which is visualized \Cref{fig:pipeline}.
\alert{First the aligner is trained on the pre-processed data and a forced alignment generated as duration reference for the \gls{TTS} model.
With that the \gls{TTS} is trained and used for generation of synthetic data.
Afterwards the \gls{ASR} models are trained on the generated data for a final evaluation.}
We remove unnatural long silence portions by using the silence filter from FFmpeg\footnote{\url{https://ffmpeg.org/ffmpeg-filters.html\#silenceremove}} with a threshold of -50dB.
Synthesis is done on two portions of data, the TTS training data and a similar amount of unseen text data.
Synthesizing the data seen during \gls{TTS} allows our analysis to be done with as little \gls{TTS} errors as possible, while using unseen data helps verifying the validity of our results.
In the baseline case we randomize the speaker ID, using speaker IDs of train-clean-100. 
The synthesized data is then used to train the different \gls{ASR} systems \textbf{without} any additional real data.
Similar to \cite{10389782}, we additionally let the \gls{TTS} model synthesize the data with access to the target durations which were seen during training. We denote this in the tables by marking the durations as \textit{real} durations.

\section{Experiments}

\subsection{Data and Training}

\begin{table}
	\caption[Train and cross-validation (CV) mean-average error (MAE) Scores of \gls{TTS} models.]{\alert{Train and cross-validation (CV) mean-average error (MAE) Scores of \gls{TTS} models.}}
	\centering
	\adjustbox{width=0.48\textwidth,keepaspectratio}{
		\begin{tabular}{| c | c | c | c | c | c |}
			\hline
			Scale & Model & \multicolumn{2}{c|}{Spectrogram MAE} & \multicolumn{2}{c|}{Duration MAE}\\
			\cline{3-6}
			Type & Scale & Train & CV & Train & CV \\
			\hline
			\hline
			- & - &0.305 & 0.376& 1.373& 1.461\\
			\hline
			\multirow{2}{*}{Dimension} & $\gamma = 1.5$ & 0.278 & 0.375 & 1.201 & 1.448 \\			
			\cline{2-6}				
			&$\gamma = 2.0$ & 0.264 & 0.373 & 1.090 & 1.453 \\
			\hline
			Layers & $\omega = 2\phantom{.0}$ & 0.296 & 0.369 & 1.333 & 1.451 \\			
			\hline			
	\end{tabular}}
	\vspace{-0.5em}
	\label{tab:maes}
\end{table}
For all our experiments we use the the train-clean-100 subset of LibriSpeech as supervised training data.
While common literature usually synthesizes an unseen part of train-clean-360 \cite{Laptev-2020-YouDoNotNeedMore, 9688255, Baskar-2021-EatEnhancedASR-TT}, we also conduct experiments on the previously seen training data.
Comparing results to synthesizing unseen data, we can observe generalization effects of the TTS. 
As \gls{CV} set for \gls{ASR} training we use a combination of dev-clean and dev-other. For \gls{TTS} training we construct our own \gls{CV} set where we split 4 sequences per speaker from the training data resulting in 1004 sequences. 
To generate phoneme representations for words not contained in the LibriSpeech lexicon we use Sequitur \cite{Bisani-2008-Joint-sequencemodel}.
For the \gls{AED} model we use byte-pair encoding \cite{sennrich-etal-2016-neural} with 2000 merge operations.
We evaluate all of our models on \textit{dev-clean} and \textit{dev-other} and do not apply silence removal.
The \gls{NAR}-\gls{TTS} is trained for 200 steps, the \gls{HMM} for 100 EM-steps, the \gls{Hybrid} for $\sim$13 full epochs and the \gls{AED} model for $\sim$165 full epochs.
All experiments were done on a single consumer 11Gb GPU for easy reproducibility. As optimizer we use Adam \cite{DBLP:journals/corr/KingmaB14} with a learning rate decay factor of 0.9 based on the CV score.

\subsection{Effect of Vocoding and Data Preprocessing}

\begin{table*}[t]
	\caption[Speaker Embedding Results]{Speaker Embedding results. WER [\%] evaluation of LibriSpeech \textit{dev-clean} and \textit{dev-other} corpora. Gaussian upsampling with SAT alignment used for \gls{TTS}. No shuffling means \gls{TTS} sees the same speaker embedding as during training.}	
	\vspace{-1.0em}
		\begin{center}
			\begin{tabular}{| c | c | c | c | c | c | c | c | c |}
				\hline
				\multirow{3}{*}{Data}& \multirow{3}{*}{\shortstack[c]{TTS- \\Durations}} & \multirow{3}{*}{\shortstack[c]{Embedding \\ Type}} & \multirow{3}{*}{\shortstack[c]{Shuffle \\ Embedding}} & \multicolumn{5}{c|}{\alert{WER} [\%]} \\
				\cline{5-9}
				& & & & \multicolumn{2}{c|}{\gls{HMM}} & \multicolumn{2}{c|}{\gls{AED}} & \gls{Hybrid} \\
				\cline{5-9}
				& &  &  & clean & other & clean & other & other\\
				\hline
				\hline
				Real & - & - & - & \textbf{\phantom{0}8.1} &\textbf{25.9} &\textbf{\phantom{0}7.5} &\textbf{18.9}&\textbf{15.0}\\
				\hline
				\hline
				\multirow{12}{*}{Synthetic} & \multirow{6}{*}{pred.} & \multirow{2}{*}{Linear}& \multirow{1}{*}{Yes} &  10.0 & \textbf{32.2} & \textbf{14.1} & 37.6 & 26.2 \\
				\cline{4-9}
				& & & No & \textbf{\phantom{0}9.8} & 33.0 & 14.7 & 39.3 & 25.5\\
				\cline{3-9}
				& &\multirow{2}{*}{X-Vectors} & Yes  & 10.3 & 33.3 & 16.1 & 40.7 & 26.4\\	
				\cline{4-9}
				& & & \multirow{1}{*}{No} & 10.3 & 32.5 & \textbf{14.1} & 36.5 & 24.0\\						
				\cline{3-9}
				& &\multirow{2}{*}{Resemblyzer} & \multirow{1}{*}{Yes} & 10.1 & 32.4 & 15.5 & 38.4 & 25.2 \\	
				\cline{4-9}
				& & & \multirow{1}{*}{No} & 10.1 & 32.6 & 14.6 & \textbf{36.2} & \textbf{23.8}\\				
				\cline{2-9}			
				&\multirow{6}{*}{real} & \multirow{2}{*}{Linear}& \multirow{1}{*}{Yes} &  \phantom{0}9.7 & 32.2 & 10.9 & 31.8 & 26.8 \\
				\cline{4-9}
				& & & No & \phantom{0}9.6 & 33.1 & 10.9 & 31.6 & 26.2 \\
				\cline{3-9}
				& & \multirow{2}{*}{X-Vectors} & Yes & \phantom{0}9.8 & 32.4 & 10.9 & 32.6 & 26.6\\		
				\cline{4-9}
				& & &\multirow{1}{*}{No} & \textbf{\phantom{0}9.5} & \textbf{31.2} & 10.1 & 28.3 & \alert{24.7}\\				
				\cline{3-9}
				& & \multirow{2}{*}{Resemblyzer} & \multirow{1}{*}{Yes} & \phantom{0}9.8 & 32.0 & 11.0 & 31.2 & 25.7\\		
				\cline{4-9}
				& & & \multirow{1}{*}{No} & \phantom{0}9.8 & \textbf{31.0} & \textbf{\phantom{0}9.9} & \textbf{27.8} & \alert{\textbf{24.5}}\\	
				\hline				
			\end{tabular}
		\end{center}
	\label{tab:speaker_emb}
	\vspace{-1.0em}
\end{table*}

\Cref{tab:baseline} shows a comparison of our three baseline models.
For this we first train the models on train-clean-100, with and without the silence removal.
All three models degrade by a similar amount, which is to be expected, since silence portions of both training and test data differ.
In lines three and four of \Cref{tab:baseline}, the effect of vocoding is shown.
For this we extract mel features from the real data and convert them back to audio with our vocoder.
Here a first difference of the ASR models becomes visible. The degradation of \gls{HMM} ranges from 0.8\% to 1.8\% absolute for dev-other, depending on the inclusion of silence removal.
For the two neural models this degradation is much less, ranging from an improvement of 0.2\% to a degradation of 0.6 \% \gls{WER}.
When replacing the real data with \gls{TTS} generated audio again the models behave differently.
For dev-clean \gls{HMM} is able to keep the best performance of $\sim$25\% relative increase, while for \gls{AED} with predicted phoneme durations the \gls{WER} doubles.
When using the aligned durations the \gls{WER} of the \gls{AED} improves by 4\% absolute while improvements for \gls{HMM} are only marginal.
For dev-other this effect is similar, but in this case dominated by the fact that the \gls{HMM} is already showing weak performance on the more noisy data.
For the \gls{Hybrid} model the performance degrades by around two-thirds, with the special exception that \gls{TTS} with oracle durations do not help the model. 
\alert{While the \gls{TTS} converges without silence removal, contrary to previous experiments for autoregressive models in \cite{9688255}, results are significantly worse and thus omitted from the table.}

\subsection{Model sizes and Generalization}

Next-up we investigate the influence of different model sizes on both the generalization capabilities of the \gls{TTS} model and the influence on \gls{ASR} training.
For this we chose two different model scaling approaches.
TTS literature usually reports on a single set of hyperparameters, but due to the possibility of automatic evaluation through ASR training and recognition, we can conduct a study on different choices.
 In the first approach we scale \gls{TTS} model dimensions by a factor $\gamma$, meaning that e.g. for $\gamma = 1.5$ the 512-dimensional layers are increased to 768.
 Analogue we indicate scaling the layer amount with $\omega$, meaning that e.g. for $\omega = 2$ there are twice as many \gls{LSTM} layers in the \gls{TTS} models. 
As seen in the upper part of \Cref{tab:general} increasing the model size or the layer count for audio generation in both cases helps the \gls{ASR} models.
Here the \gls{AED} model benefits the most from the larger models, increasing relative performance on dev-clean by $\sim$30\% and on dev-other by $\sim$20\%.
While the \gls{Hybrid} model shows improvements of $\sim$15\% on dev-other, the \gls{HMM} model only improves marginally with data generated by larger \gls{TTS} models. 
This also contradicts to the common idea that deeper models are able to hold even with larger models, while having a friction of the parameters.
In the case of generating synthetic data for \gls{ASR} training this paradigm seems to be not trivially realizable. 
We hypothesize that larger \gls{TTS} systems are able to better replicate the acoustic structure of the training data, which then reflects in better synthesis of seen data.
In the lower section of \Cref{tab:general} we show results on an unseen portion of train-clean-360.
Namely\alert{,} we chose \alert{transcriptions} that in the original corpus relates to 100h of data, hence the name 100h-LS-360.
As to be expected the general performance of the \gls{TTS} on unseen data is worse than on the training data, still the larger \gls{TTS} models are able to improve the performance of the \gls{ASR} models, even though training scores indicated overfitting, as visible in \Cref{tab:maes}.
This confirms our perception that for synthetic data generation \gls{TTS} training loss scores are not meaningful as an indicator for generalisation and performance on a held-out dataset.
A notable difference is visible in the performance of the different \gls{ASR} models on the unseen data.
The \gls{HMM} is almost able to replicate the performance compared to the seen training data, with an absolute difference of 0.3\% WER on dev-clean and dev-other for the best results.
A similar result is visible for the \gls{Hybrid} model, where the degradation is only 1.1\% WER absolute.
In the case of \gls{AED} the performance is a lot worse, where the best performance with unseen data degrades by almost 50\% relative on dev-clean and $\sim$15\% relative on dev-clean.
This indicates that modelling errors done by the \gls{TTS} model during synthesis of unseen data hurt the performance of the \gls{AED} a lot more than for the other two models. 

\subsection{Speaker Representations}

\label{sec:embedding}
As a last study we investigate the influence of the speaker embedding on the performance of the synthetic data in \gls{ASR} training.
In order to generate more expressive speaker embeddings we train an X-Vector model \cite{XVectors} on train-clean-100, as well as taking embeddings directly from the pre-trained \alert{ \textit{Resemblyzer}\footnote{\url{https://github.com/resemble-ai/Resemblyzer}} model \cite{jia2018transfer}}.
We train our TTS model by replacing the lookup table by the generated embeddings.
The results of this can be found in \Cref{tab:speaker_emb}.
For our baseline keeping the speaker tags random does not make the synthesized data worse for \gls{ASR} training.
Adding the embeddings generated by both X-Vectors and Resemblyzer does not help improve over the initial baseline and rather the performance degrades, especially in the case of AED.
Only when not shuffling the embeddings during synthesis the pre-trained embeddings are able to keep up with the baseline, surpassing it together with real durations.
From this we conclude that overall the speaker embeddings generated by the TTS model generalize well.
Feeding embeddings from more elaborate models without changes to the model makes the TTS overfit instead of benefiting from richer embeddings.
Nevertheless, in the correct setting, they can provide meaningful information to the model, as seen in the results with real durations.

\section{Conclusions}

\alert{In this work we used a text-to-speech (TTS) model for the generation of synthetic data for automatic speech recognition (ASR) training.
We modified the \gls{TTS} system in different aspects and investigated how this impacts the ASR training on the synthetically generated data.
Increasing the size of the \gls{TTS} led to more overfitting according to the training and validation scores.
Still, when using the enlarged TTS for synthetic data generation, the \gls{ASR} performance would improve.
This means hyperparameter tuning for \gls{TTS} and proper evaluation is required before drawing conclusions from \gls{ASR} training procedures involving synthetic data.
Basing model selection solely on loss scores does not suffice.
In a second set of experiments, we increased the \gls{TTS} complexity by adding pre-trained networks for speaker modeling. 
In contrast to enlarging the model, the results were less conclusive.
Only in some of the experimental settings the \gls{ASR} performance would improve, and the improvements were not consistent among the different \gls{ASR} architectures used.
It seems that the TTS tends to overfit on the given embeddings, which is reflected in the performance increase when using real durations and using the same speaker embedding as seen during training for synthesis.
We made the additional observation, that the vocoding of log-mel-features using a low-quality method such as Griffin \& Lim does not strongly degrade the utilization of the audio data for \gls{ASR} training, while reducing the overhead for data generation significantly.
Overall we have seen that gap between real and synthetic data is smaller for traditional \gls{ASR} systems.
Real phoneme variations and stronger speaker embeddings affected these systems much less than an attention-encoder-decoder \gls{ASR} systems.
Future work should aim to find suitable aspects in synthetic data which correlate with the ASR performance across different model conditions.
}

\section{Acknowledgments}
{This work was partially supported by NeuroSys, which as part of the initiative “Clusters4Future” is funded by the Federal Ministry of Education and Research BMBF (03ZU1106DA), and by the project RESCALE within the program \textit{AI Lighthouse Projects for the Environment, Climate, Nature and Resources} funded by the Federal Ministry for the Environment, Nature Conservation, Nuclear Safety and Consumer Protection (BMUV), funding ID: 67KI32006A.}

\bibliographystyle{IEEEtran}
\bibliography{mybib}

\begin{thebibliography}{10}
\providecommand{\url}[1]{#1}
\csname url@samestyle\endcsname
\providecommand{\newblock}{\relax}
\providecommand{\bibinfo}[2]{#2}
\providecommand{\BIBentrySTDinterwordspacing}{\spaceskip=0pt\relax}
\providecommand{\BIBentryALTinterwordstretchfactor}{4}
\providecommand{\BIBentryALTinterwordspacing}{\spaceskip=\fontdimen2\font plus
\BIBentryALTinterwordstretchfactor\fontdimen3\font minus
  \fontdimen4\font\relax}
\providecommand{\BIBforeignlanguage}[2]{{%
\expandafter\ifx\csname l@#1\endcsname\relax
\typeout{** WARNING: IEEEtran.bst: No hyphenation pattern has been}%
\typeout{** loaded for the language `#1'. Using the pattern for}%
\typeout{** the default language instead.}%
\else
\language=\csname l@#1\endcsname
\fi
#2}}
\providecommand{\BIBdecl}{\relax}
\BIBdecl

\bibitem{Laptev-2020-YouDoNotNeedMore}
A.~Laptev, R.~Korostik, A.~Svischev, A.~Andrusenko, I.~Medennikov, and
  S.~Rybin, ``You do not need more data: Improving end-to-end speech
  recognition by text-to-speech data augmentation,'' \emph{2020 13th
  International Congress on Image and Signal Processing, BioMedical Engineering
  and Informatics ({CISP}-{BMEI})}, October 2020.

\bibitem{9688255}
N.~Rossenbach, M.~Zeineldeen, B.~Hilmes, R.~Schlüter, and H.~Ney, ``Comparing
  the benefit of synthetic training data for various automatic speech
  recognition architectures,'' in \emph{2021 IEEE Automatic Speech Recognition
  and Understanding Workshop (ASRU)}, 2021, pp. 788--795.

\bibitem{8683480}
A.~Tjandra, S.~Sakti, and S.~Nakamura, ``End-to-end feedback loss in speech
  chain framework via straight-through estimator,'' in \emph{ICASSP 2019 - 2019
  IEEE International Conference on Acoustics, Speech and Signal Processing
  (ICASSP)}, 2019, pp. 6281--6285.

\bibitem{Baskar-2021-EatEnhancedASR-TT}
M.~K. Baskar, L.~Burget, S.~Watanabe, R.~F. Astudillo, and J.~H. Cernocky,
  ``Eat: Enhanced {ASR}-{TTS} for self-supervised speech recognition,'' in
  \emph{{ICASSP} 2021 - 2021 {IEEE} International Conference on Acoustics,
  Speech and Signal Processing ({ICASSP})}.\hskip 1em plus 0.5em minus
  0.4em\relax {IEEE}, June 2021.

\bibitem{Hu-2022-SYNTUtilizingIm}
T.-Y. {H}u, M.~{A}rmandpour, A.~{S}hrivastava, J.-H.~R. {C}hang, H.~{K}oppula,
  and O.~{T}uzel, ``{SYNT}++: {U}tilizing {I}mperfect {S}ynthetic {D}ata to
  {I}mprove {S}peech {R}ecognition,'' in \emph{{ICASSP} 2022 - 2022 {IEEE}
  {I}nternational {C}onference on {A}coustics, {S}peech and {S}ignal
  {P}rocessing ({ICASSP})}.\hskip 1em plus 0.5em minus 0.4em\relax {IEEE}, May
  2022.

\bibitem{anastassiou2024seedtts}
Seed-Team and ByteDance, ``Seed-tts: A family of high-quality versatile speech
  generation models,'' \emph{ArXiv}, vol. abs/2406.02430, p.~5, 2024.

\bibitem{bourlard2012connectionist}
H.~A. Bourlard and N.~Morgan, \emph{Connectionist speech recognition: a hybrid
  approach}.\hskip 1em plus 0.5em minus 0.4em\relax Springer Science \&
  Business Media, 2012, vol. 247.

\bibitem{7472621}
W.~Chan, N.~Jaitly, Q.~Le, and O.~Vinyals, ``Listen, attend and spell: A neural
  network for large vocabulary conversational speech recognition,'' in
  \emph{2016 IEEE International Conference on Acoustics, Speech and Signal
  Processing (ICASSP)}, 2016, pp. 4960--4964.

\bibitem{Shen-2018-NaturalTTSSynthesi}
J.~Shen, R.~Pang, R.~J. Weiss, M.~Schuster, N.~Jaitly, Z.~Yang, Z.~Chen,
  Y.~Zhang, Y.~Wang, R.~Skerrv-Ryan, R.~A. Saurous, Y.~Agiomvrgiannakis, and
  Y.~Wu, ``Natural {TTS} synthesis by conditioning wavenet on {MEL} spectrogram
  predictions,'' in \emph{{ICASSP} 2018 - 2018 {IEEE} International Conference
  on Acoustics, Speech and Signal Processing ({ICASSP})}.\hskip 1em plus 0.5em
  minus 0.4em\relax {IEEE}, April 2018.

\bibitem{9688218}
S.~Ueno, M.~Mimura, S.~Sakai, and T.~Kawahara, ``Data augmentation for asr
  using tts via a discrete representation,'' in \emph{2021 IEEE Automatic
  Speech Recognition and Understanding Workshop (ASRU)}, December 2021, pp.
  68--75.

\bibitem{10389782}
N.~Rossenbach, B.~Hilmes, and R.~Schlüter, ``On the relevance of phoneme
  duration variability of synthesized training data for automatic speech
  recognition,'' in \emph{IEEE Automatic Speech Recognition and Understanding
  Workshop (ASRU)}, 2023, pp. 1--8.

\bibitem{Perez-Gonzalez-de-Martos-2021-VRAIN-UPVMLLPssys}
\BIBentryALTinterwordspacing
A.~{P}érez-{G}onzález-de {M}artos, A.~{S}anchis, and A.~{J}uan,
  ``{VRAIN}-{UPV} {MLLP}'s system for the {B}lizzard {C}hallenge 2021,''
  \emph{{F}estvox {B}lizzard {C}hallenge 2021}, October 2021. [Online].
  Available: \url{http://arxiv.org/abs/2110.15792v1}
\BIBentrySTDinterwordspacing

\bibitem{Shen-2020-Non-AttentiveTacotr}
J.~Shen, Y.~Jia, M.~Chrzanowski, Y.~Zhang, I.~Elias, H.~Zen, and Y.~Wu,
  ``Non-attentive tacotron: Robust and controllable neural {TTS} synthesis
  including unsupervised duration modeling,'' \emph{ArXiv}, vol.
  abs/2010.04301v3, October 2020.

\bibitem{DBLP:conf/icassp/GriffinDL84}
\BIBentryALTinterwordspacing
D.~W. Griffin, D.~S. Deadrick, and J.~S. Lim, ``Speech synthesis from
  short-time fourier transform magnitude and its application to speech
  processing,'' in \emph{{ICASSP} '84, San Diego, California, USA, March 19-21,
  1984}, pp. 61--64. [Online]. Available:
  \url{https://doi.org/10.1109/ICASSP.1984.1172423}
\BIBentrySTDinterwordspacing

\bibitem{Bisani-2008-Joint-sequencemodel}
M.~{B}isani and H.~{N}ey, ``{J}oint-sequence models for grapheme-to-phoneme
  conversion,'' \emph{{S}peech {C}ommunication}, vol.~50, no.~5, pp. 434--451,
  May 2008.

\bibitem{Wiesler-2014-RASRNNTheRWTHne}
S.~Wiesler, A.~Richard, P.~Golik, R.~Schl\"uter, and H.~Ney, ``{RASR}/{NN}: The
  {RWTH} neural network toolkit for speech recognition,'' in \emph{{ICASSP}
  2014 - 2014 {IEEE} International Conference on Acoustics, Speech and Signal
  Processing ({ICASSP})}.\hskip 1em plus 0.5em minus 0.4em\relax {IEEE}, May
  2014.

\bibitem{Ren-2020-FastSpeech2Fasta}
Y.~{R}en, C.~{H}u, X.~{T}an, T.~{Q}in, S.~{Z}hao, Z.~{Z}hao, and T.-Y. {L}iu,
  ``{F}ast{S}peech 2: {F}ast and {H}igh-{Q}uality {E}nd-to-{E}nd {T}ext to
  {S}peech,'' in \emph{{I}nternational {C}onference on {L}earning
  {R}epresentations ({ICLR})}, December 2021.

\bibitem{librispeech}
V.~Panayotov, G.~Chen, D.~Povey, and S.~Khudanpur, ``Librispeech: An asr corpus
  based on public domain audio books.'' in \emph{ICASSP}.\hskip 1em plus 0.5em
  minus 0.4em\relax IEEE, 2015, pp. 5206--5210.

\bibitem{CART}
K.~Beulen and H.~Ney, ``Automatic question generation for decision tree based
  state tying,'' in \emph{Proceedings of the 1998 IEEE International Conference
  on Acoustics, Speech and Signal Processing, ICASSP '98 (Cat. No.98CH36181)},
  vol.~2, May 1998, pp. 805--808.

\bibitem{park19e_interspeech}
D.~S. Park, W.~Chan, Y.~Zhang, C.-C. Chiu, B.~Zoph, E.~D. Cubuk, and Q.~V. Le,
  ``{SpecAugment: A Simple Data Augmentation Method for Automatic Speech
  Recognition},'' in \emph{Proc. Interspeech 2019}, 2019, pp. 2613--2617.

\bibitem{sennrich-etal-2016-neural}
R.~Sennrich, B.~Haddow, and A.~Birch, ``Neural machine translation of rare
  words with subword units,'' in \emph{Proceedings of the 54th Annual Meeting
  of the Association for Computational Linguistics (Volume 1: Long
  Papers)}.\hskip 1em plus 0.5em minus 0.4em\relax Association for
  Computational Linguistics, Aug. 2016, pp. 1715--1725.

\bibitem{DBLP:journals/corr/KingmaB14}
D.~P. Kingma and J.~Ba, ``Adam: {A} method for stochastic optimization,'' in
  \emph{3rd International Conference on Learning Representations, {ICLR} 2015,
  San Diego, CA, USA, May 7-9, 2015, Conference Track Proceedings}, Y.~Bengio
  and Y.~LeCun, Eds., 2015.

\bibitem{XVectors}
D.~Snyder, D.~Garcia-Romero, G.~Sell, D.~Povey, and S.~Khudanpur, ``X-vectors:
  Robust dnn embeddings for speaker recognition,'' in \emph{2018 IEEE
  International Conference on Acoustics, Speech and Signal Processing
  (ICASSP)}, April 2018, pp. 5329--5333.

\bibitem{jia2018transfer}
Y.~Jia, Y.~Zhang, R.~Weiss, Q.~Wang, J.~Shen, F.~Ren, P.~Nguyen, R.~Pang,
  I.~Lopez~Moreno, Y.~Wu \emph{et~al.}, ``Transfer learning from speaker
  verification to multispeaker text-to-speech synthesis,'' \emph{Advances in
  neural information processing systems}, vol.~31, 2018.

\end{thebibliography}

\end{document}